\documentclass[runningheads]{llncs}
\usepackage{graphicx}
\usepackage{amsfonts, amsmath}
\usepackage[show]{chato-notes}
\begin{document}
\title{Statistics and explainability: a fruitful alliance}
%
%
\author{Valentina Ghidini}
\authorrunning{V. Ghidini}
%
\institute{Euler Institute, Universit\`a della Svizzera Italiana\\
Lugano, Switzerland\\
\email{valentina.ghidini@usi.ch}}
\maketitle              
\begin{abstract}

In this paper, we propose standard statistical tools as a solution to commonly highlighted problems in the explainability literature. Indeed, leveraging statistical estimators allows for a proper definition of explanations, enabling theoretical guarantees and the formulation of evaluation metrics to quantitatively assess the quality of explanations. This approach circumvents, among other things, the subjective human assessment currently prevalent in the literature. Moreover, we argue that uncertainty quantification is essential for providing robust and trustworthy explanations, and it can be achieved in this framework through classical statistical procedures such as the bootstrap. However, it is crucial to note that while Statistics offers valuable contributions, it is not a panacea for resolving all the challenges. Future research avenues could focus on open problems, such as defining a purpose for the explanations or establishing a statistical framework for counterfactual or adversarial scenarios.


\keywords{Explainability  \and Statistics.}
\end{abstract}

\section{Introduction}

This paper takes an unconventional approach, for either computer scientists or statisticians. It will not delve into technicalities, present groundbreaking methodologies or introduce new explainability techniques. It will however explore a reflection on the use of statistics in the explainability context to argue that a closer relationship between standard statistical procedures and explainability techniques is needed to tackle some foundational challenges.

Considerable debate has revolved around the desirability of employing black-box models, especially in highly sensitive applications~\cite{Rudin2019}. However, in this work, we acknowledge the impracticality of completely eliminating the use of black-box models in real-world applications. Consequently, we start from the assumption that there is a necessity for explainability techniques, recognizing the need to provide explanations for black boxes somehow, since such inscrutable models are not going to disappear anytime soon. 

The main question at the core of this work is whether we can address certain challenges currently undermining the foundations of eXplainable Artificial Intelligence (XAI) through the application of statistical methods, analyzing the corresponding advantages and disadvantages. The aim is to persuade practitioners that statistics can serve as a valuable tool, offering a robust ground for the introduction of novel methods and techniques. Specifically, we try to leverage statistics to systematically tackle existing challenges in the realm of explainability, employing a formal mathematical methodology to address these open issues.

\subsection{Related literature}

Numerous works in the literature have brought attention to foundational issues and ongoing challenges in XAI, but, to our knowledge, no tentative solution has been outlined yet. This paper draws inspiration from various sources, including but not limited to~\cite{lipton2017,Guidotti2018,Rudin2019,Miller2019,gunning2019,rudin2021,rawal2022,freiesleben2023}.
As previously mentioned, our objective is to show how statistics can alleviate (if not entirely resolve) various complaints and challenges posed by researchers. Standard statistical procedures can indeed offer partial or comprehensive solutions to the principal issues, as outlined in a detailed point-by-point analysis in the next sections.

The remainder of this paper is structured as follows: Section~\ref{sec:issues} lists some of the issues currently contained in the literature that are the target of our analysis. Section~\ref{sec:solution} addresses such issues point-by-point, illustrating how statistical procedure can be used. Section~\ref{sec:perks} highlights some additional perks of using statistical explanations, while Section~\ref{sec:open_problems} addresses some problems which can not be solved merely applying standard statistical procedures. Then, Section~\ref{sec:discussion} contains some conclusive thoughts about future and desirable research directions.

\section{Foundational issues in explainability}\label{sec:issues}

In this section, we enumerate foundational issues in explainability as identified in various works within the literature. In particular, we list the challenges we aim to tackle using standard statistical procedures.

\subsection{Issue 1: lack of proper definitions}\label{subsec:issue_definition}

First, let us address the biggest elephant in the room: the absence of a clear, unique and precise definition for the term \textit{explanation}. Numerous scholars have already pointed out this issue in works such as~\cite{salmon1984,Guidotti2018,Miller2019,buijsman2022} and references therein. While certain referenced works offer definitions of the term \textit{explanation}, these definitions often lack practical applicability and, as a result, they are not {actionable}. In other words, translating the English meaning into a precise mathematical definition of explanations is usually difficult, if not impossible.
For example, citing~\cite{Guidotti2018}, the authors define an explanation as ``\textit{an ``interface''
between humans and a decision maker that is at the same time both an accurate
proxy of the decision maker and comprehensible to humans}''. However, such definition is still ambiguous, as it remains unclear what, e.g., an interface is from either a mathematical or an algorithmic viewpoint, or how to obtain one. 
Another definition of explanation provided in~\cite{ciatto2020} is the following: ``\textit{we define the act of ``interpreting'' some object $X$ as the activity performed by an agent $A$ ... assigning a subjective meaning to $X$. Such meaning is what we call interpretation. We define “explaining” as the activity of producing a more interpretable object $X'$ out of a less interpretable one, namely $X$, performed by agent $A$}''. While this definition introduces a notion reminiscent of algebraic concepts, it still falls short in providing concrete information about what an explanation is, its form, or the methodology for obtaining one. 
As of our current understanding, it appears that there is a recurrent, ephemeral illusion in the literature, according to which the English definition of explanation is perceived to seamlessly convert into an actionable technical definition, which, in reality, is often not the case. The lack of a proper definition for explanations is one of the root causes of the issues discussed in the next subsections, namely the lack of theoretical guarantees, simple quantitative assessments of the goodness of explanations, and uncertainty quantification.

Arguably, a class of more actionable explanations implicitly defines what in statistics is known as variable importance measures. For example, the work in~\cite{das2020} gives the following definition: ``\textit{An explanation is additional meta information, generated by an external algorithm or by the machine learning model itself, to describe the feature importance or relevance of an input instance towards a particular output classification}''. Even though not explicitly mentioned, it is natural for statisticians to think of variable importance measures, such as the world-famous ones introduced by Breiman in~\cite{Breiman2001}. In the next section, we argue that, among other things, statistics can be seen as the bridging gap enabling the interpretation of traditional variable importance measures as explanations for black-box models.

\subsection{Issue 2: lack of theoretical guarantees}
Another problem is the lack of theoretical guarantees about the explanations. Consider standard XAI approaches, such as GradCAM~\cite{Selvaraju2020}, LIME~\cite{Ribeiro2016}, RISE~\cite{Petsiuk2018} among others. These works offer empirical algorithms that yield explanations. Their appeal lies in the ease of interpreting the explanations and the overall sensibility of the entire procedure, contributing to their popularity and widespread adoption.
However, these approaches usually lack a theoretical analysis. For instance, when considering a probabilistic modeling framework where data are treated as realizations of random variables living on probability spaces, it remains unclear whether explanations exist in general as well-defined quantities (\textit{existence result}) and, if so, how they can be precisely defined in a mathematical sense.
Moreover, as of our current understanding, it remains uncertain whether the algorithms provide correct explanations, at least in some defined sense (\textit{correctness result}). Essentially, we argue that there is a need to precisely determine the quantity being estimated as explanation for the model. Furthermore, a desirable property of XAI algorithms would be that the estimation of the explanations improves as the number of collected data instances grows (\textit{convergence result}). Notice however that these theoretical properties can only be verified if and only if we can establish a proper mathematical definition of explanations.
All these results are standard in statistical analysis and we think they would be significant for the theoretical understanding and enhancement of current XAI techniques.

\subsection{Issue 3: lack of (simple) evaluating metrics}

A significant challenge in explainability lies in the absence of simple, quantitative evaluation metrics to assess the goodness of explanations~\cite{ghidini2019}. This difficulty is partially due to the lack of a clear and actionable definition of explanation, as previously mentioned in Subsection~\ref{subsec:issue_definition}. This makes it difficult to establish a ground truth with respect to which the evaluation of explanations can be conducted. Indeed, if the quantity defined as explanation is ill-defined or ambiguous, assessing the correctness of the output from a proposed technique is impossible.
Furthermore, the absence of a clear and actionable definition for explanations makes it difficult to establish benchmarks where the true explanation is analytically known.
Consequently, defining simulated scenarios to assess the correctness of XAI algorithms (i.e.\ whether they return the correct explanations) is unfeasible. 


Since the lack of precise quantitative metrics inhibits a quantitative and objective evaluation, the literature often resorts to sanity checks performed by humans, as highlighted by~\cite{freiesleben2023}. These assessments however come at a cost --- they are time-consuming, require financial resources, and are subject to individual judgment.  In fact, in this context, explanations are often considered correct if and only if they align with human judgment regarding the importance of features. For instance, in a cat-dog image classifier, an explanation is deemed correct if it identifies the pixels representing the actual animal as the most crucial part of the input image according to the model. However, this enables a vicious cycle: certain explanations may be considered inadequate or, worse, incorrect simply because they do not align with human expectations. Nevertheless, it is plausible that the machine learning model in question operates in an unexpected manner, and explanations could be valuable in detecting discrepancies between the human symbolic representation of features and the models' sub-symbolic system.


\subsection{Issue 4: lack of uncertainty and robustness evaluation}

Last but not least, one crucial aspect often missing in many XAI techniques is the uncertainty quantification of the resulting explanations. Relying solely on point estimates of the explanations is not the best practice; it is essential to gain an understanding of how much these estimates vary across different data instances or realizations of the training set. Bayesian statisticians are well aware of this, and they typically provide estimates of the entire posterior distribution for the quantities they are interested in inferring.
Nevertheless, uncertainty quantification is not yet a common practice in the XAI community. Generally, only single-shot outputs of explanatory algorithms are provided, offering no insight into the robustness of the proposed explanations. The variability of the proposed explanations is usually not thoroughly studied or understood. For instance, it is often unclear how the explanations would change with slight alterations to the model, data points, or when using a different, randomly selected dataset. If the explanations exhibit significant variability, one may hesitate to trust them, or alternatively, one might consider collecting additional data for a more reliable estimation.

\section{A (tentative) answer: Statistics}\label{sec:solution}

In this section, we aim to offer point-by-point solutions to the challenges outlined in Section~\ref{sec:issues}. We argue that standard statistical tools and techniques can be leveraged to sketch several solutions. 

\subsection{Solution 1: mathematical measures and definitions}
 First, we can define explanations in a statistical sense, and in this framework the most natural choice is to employ the notion of variable importance measures, \textcolor{black}{as in the seminal works in}~\cite{Breiman1984,Breiman2001}.  
 \textcolor{black}{For instance}, given a set of explanatory variables~$X = (X_1,\dots, X_p)$, we can define explanations as expected value of variations in some output of the model~$f$ with respect to changes in a specific variable $X_j$ (possibly multidimensional). An easy, intuitive and general approach is to define a difference~$d(\cdot, \cdot)$ of summary statistics~$s_f(\cdot)$ pertaining to the model of interest and consider its expected value~$\mathbb{E}_{{X_{\text{interest}}}} [d\big( s_f(X_j), s_f(\widetilde{X}_j) \big)]$, where $\widetilde{X}_j$ is a perturbed version of~$X_j$\textcolor{black}{. Classical examples of perturbations of inputs are summation of white noise~\cite{Petsiuk2018}, or random permutation of the values~\cite{Breiman1984}.}
 The expected value is taken with respect to ${X_{\text{interest}}} = X_j$ (as in sensitivity analysis~\cite{borgonovo2023}) or ${X_{\text{interest}}} = X_{(-j)}$, that is with respect to all the other covariates. 
 To summarize, \textcolor{black}{in this example}, the explanations are defined as expected values of differences in summary statistics of the model outputs, whenever $X_j$ is perturbed. This formulation yields explanations which can be written as:
\begin{equation}\label{eq:importance}
    I_f(X_j) = \mathbb{E}_{{X_{\text{interest}}}} [d\big( s_f(X_j), s_f(\widetilde{X}_j) \big)].
\end{equation}
Given such expected values, explanations can be estimated by means of proper estimators, endowed with estimates~\cite{borgonovo2023,ghidini2023_xai}.
As previously mentioned, this approach is not novel: it aligns precisely with the concept of variable importance measures. In the end, both procedures assign a score to each covariate (possibly multidimensional) which is proportional to its importance for the model. Even though variable importance measures have been around for decades, their flexibility is endless: according to the choice of $d(\cdot, \cdot)$ and of $s_f(\cdot)$ we can measure the impact of each covariate (or each group of explanatory variables) in infinite ways. 
It is then clear that such definition bridges the gap between explanations and variable importance measures, exploiting a proper statistical quantity, which can be endowed with an estimator, as explanation. 

\textcolor{black}{With this example}, we have delved into the natural interpretation of statistical variable importance measures as a general way of providing explanations which can be endowed with proper statistical estimators. However, the possibilities are endless: one can use statistical quantities such as variance, quantiles, probability distributions computed either on differences on the data or the model outputs to define explanations suited for the application at hand.
\textcolor{black}{Indeed, notice that explanations defined as in Equation~\eqref{eq:importance} are just one (flexible) possibility showcased as example. In general, the aim of this subsection is to advocate for providing explanations using statistical quantities to accurately capture, through their definition, specific characteristics of the model to be explained. Another example of well-known tools in XAI are partial dependence plots~\cite{goldstein2013} or accumulated local effects plots~\cite{Apley2020}. In both cases, the explanations are provided by effects or measures of dependences defined through probabilistic expected values. As such, those kind of explanations (and many others) can be analyzed with a statistical viewpoint in mind, employing all the techniques discussed in the next sections to obtain proper statistical explanations.}

\subsection{Solution 2: studies of convergence as theoretical guarantees}

With a proper statistical definition of explanation in place, it is natural to leverage standard statistical tools to establish convergence results. For example, consider the previously defined explanations $I_f(X_j) $ \textcolor{black}{as a running example} and suppose we observe $n$ independent and identically distributed $p-$dimensional data points $\mathbf{x}_1, \dots, \mathbf{x}_n$, which are commonly arranged row-wise into a design matrix $X$. 
Then, it is natural to estimate the explanations $I_f(X_j)$ using statistical estimators, such as the empirical mean. By employing such a procedure and assuming standard regularity constraints on the summary statistic $s_f(\cdot)$ and the difference $d(\cdot, \cdot)$, it is possible to leverage well-known theorems like the law of large numbers and the central limit theorem to establish convergence results. 
\textcolor{black}{The Xi~method~\cite{borgonovo2023,ghidini2023_xai} provides an example illustrating such a pipeline.} The referenced works offer theoretical convergence results for the explanations, demonstrating that as the number of data points increases, the estimated explanations converge to their true, existing values. Future endeavors could also address distributional results: it would be interesting to prove the asymptotic Gaussianity of the estimators of the explanations. This could be utilized to construct proper asymptotic confidence intervals.

\subsection{Solution 3: definition of evaluating metrics}

If the explanations are well-defined \textcolor{black}{to sensibly capture interesting features of the model under scrutiny}, it becomes straightforward to establish appropriate and quantitative evaluation metrics.
Indeed, it is easier to evaluate the disparity between the estimated explanations and the true ones. A natural metric is a simple difference between the true values of the explanations and the estimated ones in, e.g., simulated data, but various other discrepancy measures can also be employed.
Indeed, besides the quantitative assessment, it is now easy to simulate benchmarks and scenarios where the true explanations are designed. This allows for the evaluation of whether a proposed XAI technique effectively captures the true explanations underlying the data-generating process.
This approach completely eradicates the need for human studies, as the evaluation is now entirely objective, quantitative, and conducted by means of a properly defined metric. It also avoids the subjectivity associated with determining what constitutes a good explanation based on different users' perspectives, as human evaluations are inherently influenced by varying levels of expertise, education, and other factors.
Furtherly, this approach clarifies and eliminates any ambiguity regarding what constitutes a wrong explanation: an explanation (or its estimator/estimate) is considered wrong if and only if it fails to converge or is not ``close enough'' (in the newly-defined metric sense) to the true value. In such cases, one should reconsider either the algorithm or the estimator used in the procedure.

\textcolor{black}{Notice however that the previous discussion assumes that a good explanation, at least in this statistical context, is one that is estimated correctly with respect to the true quantity underlying the data. However, since interpretability is closely bounded to the human perception of data, one could also argue that a good explanation is a \textit{useful} one, that is it can be easily employed by the end users. Indeed, explanations are of little use if they are too complex to be understood or too simple to be faithful to the black box. Evaluation metrics in this regard are trickier to define, and may involve a human in the loop at some point, but more objective and quantitative alternatives employing statistical techniques could be explored in future research avenues.}


\subsection{Solution 4: uncertainty quantification}

A significant challenge is the absence of a comprehensive framework for uncertainty quantification in explanations. In general, few papers discuss the robustness of the outputs generated by proposed XAI techniques~\cite{borgonovo2023,ghidini2023_xai}. We argue that the gold standard should be to provide uncertainty quantification for the explanations, but as of the current state of the field at the time of writing, this is not yet a common practice.  Indeed, reliability and robustness of explanations are always among the essential requirements for the deployment in real-world scenarios. Intuitively, when either the data instances to be explained or the prediction process provided by the model are slightly perturbed, the explanations should not undergo significant changes.
By incorporating statistics into the process, we gain access to standard tools that can be used to provide not only point estimates of explanations but also confidence intervals, whether asymptotical or empirical. 
In cases where we have an asymptotic distribution of the estimators for the explanations, obtaining confidence intervals becomes a relatively straightforward process. For example, if one can establish that the explanations are asymptotically normal, it becomes possible to derive closed-form asymptotic confidence intervals. Even if verifying asymptotic normality is not feasible, one can resort to standard statistical resampling techniques, such as the bootstrap. In this context, the data are resampled $B$ times to obtain the $B$ different estimates of the explanations allowing for the computation of the corresponding empirical distribution, variability measures, and/or confidence interval. 
\textcolor{black}{As a final note, notice that this section only analyzes explanation uncertainty, but there exist many other sources of variability, such as model or sampling uncertainty. However, we argue that model uncertainty should be addressed prior to explaining it (in the definition/training part of the pipeline). As for sampling uncertainty, it should be analyzed in the algorithmic part: for example, if the estimation of the explanations involves some Monte Carlo algorithm, then a proper diagnostic procedure should be carried out, computing effective sample sizes and examining suitable trace plots.}

\section{Other statistical perks}\label{sec:perks}
 
There are additional significant benefits to utilizing statistics to define explanations. We list some advantages below, but such list is not comprehensive and could be updated in the future.
 
\subsection{Bonus 1: trustworthy explanations}

Using a coherent procedure to have well-defined explanations has the undeniable advantage of having assurances about the accuracy of the quantities that we are estimating. If convergence results can also be established, there are theoretical guarantees regarding the correctness of the explanation estimates when the number of data points is sufficiently large. This establishes a solid foundation for \textit{trust} in the provided explanations.

Consequently, there are two potential outcomes in the analysis of explanations: first, the explanations of the black-box models align with our human expectations, highlighting the features in the prediction process that we expect. 
In this case, we can infer that the model of interest is effectively capturing expected meaningful dependence structures in the data.
In a second scenario, the explanations may emphasize features that we, as human beings, do not consider important. In some current studies, such explanations are categorically considered incorrect. However, with theoretical assurances regarding the correctness of the estimated explanations, a more nuanced interpretation can be gained. Generally speaking, a mismatch between the most important features according to the explanations and the researcher's expectations indicates a divergence in the black-box sub-symbolic representation of the data from what was anticipated. 
Hence, it is plausible that the researcher's perception of the importance of certain features is misplaced, or there may be multicollinearity issues causing the model to favor other variables. In any case, the explanations can also serve as a diagnostic tool for the black-box model.

\subsection{Bonus 2: \textcolor{black}{statistics offers interpretable models}}

In this section, we draw inspiration from prominent works that assert post~hoc interpretability is not always the optimal choice and advocate against the use of black-box models in high-stakes decisions~\cite{Rudin2019,rudin2021}. Much of the support for these arguments stems from the observation that some XAI procedures generate explanations that lack stability, trustworthiness, and robustness. We argue that, by incorporating statistical methods, many of these issues can be alleviated, if not entirely resolved, as discussed in the previous sections.

Even though we believe black-box models will not fall in disgrace anytime soon, we acknowledge that in certain sensitive contexts, like medical or financial applications, it is preferable to use transparent models. A straightforward approach to define interpretable models is to resort to classic inferential statistics. For instance, one can utilize linear regression, generalized linear models, generalized additive models, trees, survival models, and so forth. These models are inherently interpretable and well-performing, if defined correctly. The main drawback, however, is that they necessitate a deeper understanding of statistics, as well as more meticulous prior design and feature engineering, which could be challenging or require a higher level of domain expertise. \textcolor{black}{Notice that we are not claiming that \textit{all} statistical models are interpretable: for example, nonparametric models may bee too complex to be white boxes. Nonetheless, classical statistics provides a multitude of choices to model various types of data using inherently interpretable models.}

\subsection{Bonus 3: easy-to-define fairness}

Statistics also provides the means to comprehend and rigorously assess fairness in black-box models. One straightforward approach involves introducing tests for explanations concerning sensitive attributes in the design matrix, such as gender or age. Indeed, a natural procedure for a statistician would be to define a statistical test where the null hypothesis examines the significance of an explanation, e.g.\ a variable importance measure, of a sensitive attribute being null. This can be achieved by leveraging asymptotic and distributional results of estimators of explanations. Of course, this represents the more straightforward option; alternatively, one could test differences in certain outputs of the model before and after perturbing specific sensitive factors, offering actual p-values to evaluate the fairness of the model. \textcolor{black}{Such methods are fairly naive and limited, but they serve just as a starting point for a statistical definition of fairness: future research avenues should be concerned with more sophisticated statistical structures to capture complex characterization of fairness.}

\subsection{Bonus 4: explanations for the data generating process}
\textcolor{black}{
A common topic of discussion in the field of explainability is whether explanations should solely focus on black boxes or also encompass the data generating process~(DGP)~\cite{freiesleben2023,ghidini2023_xai}. It is often argued that explanations concerning the DGP can establish the ground truth for their counterpart interpreting the black box of interest~\cite{molnar2023}. 
Although we may not entirely endorse this assertion, we acknowledge the importance of understanding the data generating process first and the black box model second. Analyzing the DGP beforehand can be crucial for identifying potential disparities between the explanations of the DGP and those of the model. 
Some works in the literature have defined pre hoc explanations~\cite{borgonovo2023,ghidini2023_xai}, that are explanations specifically devised for the data generating process of interest.
Future research endeavors may explore various techniques to establish proper definitions for pre hoc explanations.
}
\section{Open problems not solved by Statistics}\label{sec:open_problems}

In this section, we list some issues highlighted by works in the literature~\cite{freiesleben2023}, for which it is currently unclear how to address them using statistics.

\subsection{Definition of a purpose}

The work in~\cite{freiesleben2023} advocates for the establishment of a purpose in the development of an explainability technique. This aspect cannot be resolved by statistics alone, as the purpose typically varies depending on the field and heavily relies on the specific application of interest. Is the purpose to ensure a fair model? Or to verify whether the model behaves as expected? Or perhaps to examine the joint dependence structure of the covariates on the target? Or even to increase the trust in the model in practitioners?

\subsection{Simplicity of explanations}

Some authors~\cite{Guidotti2018,Miller2019} also argue for the simplicity of explanations to guarantee the usability of XAI techniques for a broad audience of users. This notion is already prevalent in the literature. An illustrative example of simple explanations is provided by heatmaps explaining models for images: it is intuitively clear that highlighted pixels in the pictures are also the most crucial for the black-box model's explanation. Generally, it is not that immediate to convey such intuitive explanations using statistical measures of differences in the output, or other statistical quantities. Additional work should ensure that explanations are easily comprehensible to a diverse audience.

\subsection{Need of researchers trained in statistics}

Another unavoidable drawback of using statistics to formulate explainability techniques is that one necessitates statistical training to appropriately define and operate in such a context. While gaining proficiency in statistics is fairly accessible for individuals with a scientific background (e.g., computer scientists, engineers, etc.), it still requires time and dedication to delve into a (potentially) new literature and engage with a distinct community, albeit not significantly distant from the ones from which XAI stemmed as a research field.

\section{Discussion}\label{sec:discussion}

In this paper, we have outlined some foundational issues of explainability highlighted in the literature by different authors.
We focus on the absence of proper definitions for explanations and of theoretical guarantees, on the challenges in defining evaluation metrics, and on the doubts regarding whether the proposed explanations are robust or exhibit excessive variability, thereby raising concerns about their trustworthiness.
We argue that a solution to all these issues can be found through the application of statistical techniques. By utilizing statistical estimators for explanations, we can leverage well-established results such as the law of large numbers or the central limit theorem to furnish theoretical guarantees for explanations.
Furthermore, it is natural to incorporate uncertainty quantification into the resulting estimates, as is customary in any statistical analysis. Statistical explanations may also enhance the trust in either the models or the outcome of the XAI procedures. 
It is also natural for a statistician to employ hypothesis testing to assess the fairness of a model. However, there is still much work to be done in this regard: statistical explanations should be comprehensible to users with various backgrounds, not necessarily statisticians by training. Additionally, each technique should be crafted with a specific purpose in mind. Future research should also define the statistical counterparts of adversarial and counterfactual explanations, which is neither straightforward nor immediate. This considerations can have implications for adversarial attacks and generative models as well.




%
%
%

\subsection*{Acknowledgement} Valentina Ghidini is funded by the SNSF starting grant ``Multiresolution methods for unstructured data'' (TMSGI2\_211684). 
\subsection*{Disclosure of Interests}
The authors have no competing interests to declare that are relevant to the content of this article. 

\bibliographystyle{splncs04}
\bibliography{biblio_xai2024}
%

\end{document}